%% file: main.tex
\documentclass[10pt,twocolumn,letterpaper]{article}

\PassOptionsToPackage{table}{xcolor}
\usepackage{iccv}              
\usepackage{multirow}
\usepackage{xcolor}


\definecolor{iccvblue}{rgb}{0.21,0.49,0.74}
\usepackage[pagebackref,breaklinks,colorlinks,allcolors=iccvblue]{hyperref}

\def\paperID{7282} 
\def\confName{ICCV}
\def\confYear{2025}

\title{DualX-VSR: Dual Axial Spatial$\times$Temporal Transformer for Real-World Video Super-Resolution without Motion Compensation}

\author{
Shuo Cao$^{1,2\S}$\thanks{Equal contribution.\quad $\dagger$ Corresponding author. \quad \S This work was done during his internship at Shanghai Artificial Intelligence Laboratory.} \quad Yihao Liu$^{2}$\textsuperscript{\small\textasteriskcentered} \quad Xiaohui Li$^{3,2}$ \quad Yuanting Gao$^{4,2}$  \quad Yu Zhou$^{5}$ \quad 
Chao Dong$^{6,2\textsuperscript{$\dagger$}}$\\
$^1$University of Science and Technology of China \quad $^2$Shanghai Artificial Intelligence Laboratory\\ \quad $^3$Shanghai Jiao Tong University \quad $^4$Tsinghua University \quad $^5$ Sun Yat-sen University \\ 
\quad $^6$Shenzhen Institute of Advanced Technology, Chinese Academy of Sciences \\
{\tt\small caoshuo@pjlab.org.cn} \quad {\tt\small liuyihao14@mails.ucas.ac.cn} \quad {\tt\small chao.dong@siat.ac.cn}
}

\begin{document}

\maketitle
\input{sec/total} 
{
    \small

\input{main.bbl}
}

\clearpage
\appendix
\renewcommand{\thesection}{\Alph{section}}  
\section*{Appendix}
\input{sec/supp}

\end{document}

%% file: sec/total.tex
\begin{abstract}

Transformer-based models like ViViT and TimeSformer have advanced video understanding by effectively modeling spatiotemporal dependencies. Recent video generation models, such as Sora and Vidu, further highlight the power of transformers in long-range feature extraction and holistic spatiotemporal modeling. However, directly applying these models to real-world video super-resolution (VSR) is challenging, as VSR demands pixel-level precision, which can be compromised by tokenization and sequential attention mechanisms. While recent transformer-based VSR models attempt to address these issues using smaller patches and local attention, they still face limitations such as restricted receptive fields and dependence on optical flow-based alignment, which can introduce inaccuracies in real-world settings. To overcome these issues, we propose Dual Axial Spatial$\times$Temporal Transformer for Real-World Video
Super-Resolution (\textbf{DualX-VSR}), which introduces a novel dual axial spatial$\times$temporal attention mechanism that integrates spatial and temporal information along orthogonal directions. DualX-VSR eliminates the need for motion compensation, offering a simplified structure that provides a cohesive representation of spatiotemporal information. As a result, DualX-VSR achieves high fidelity and superior performance in real-world VSR task.

\end{abstract}

\section{Introduction}
\label{sec:intro}

The introduction of the video transformer architecture, like ViViT~\cite{vivit} and TimeSformer~\cite{timesformer}, marked a significant breakthrough in video understanding, revealing the potential of Transformer-based models for handling spatiotemporal dependencies. Building on this, recent video generation models like Sora~\cite{sora}, Latte~\cite{latte}, and VDT~\cite{vdt} -- based on the Diffusion Transformer (DiT~\cite{dit}) architecture -- have demonstrated the strengths of Transformers in long-range modeling, feature extraction, and scalability. Notably, Sora emphasized the importance of treating video as a cohesive unit rather than isolated frames, advocating for holistic spatiotemporal modeling. This progress leads us to explore whether the successes of those straightforward video transformer architectures in high-level tasks could translate effectively to low-level video tasks, such as real-world video super-resolution (VSR).

\begin{figure}[t]
  \centering
    \includegraphics[width=0.95\linewidth]{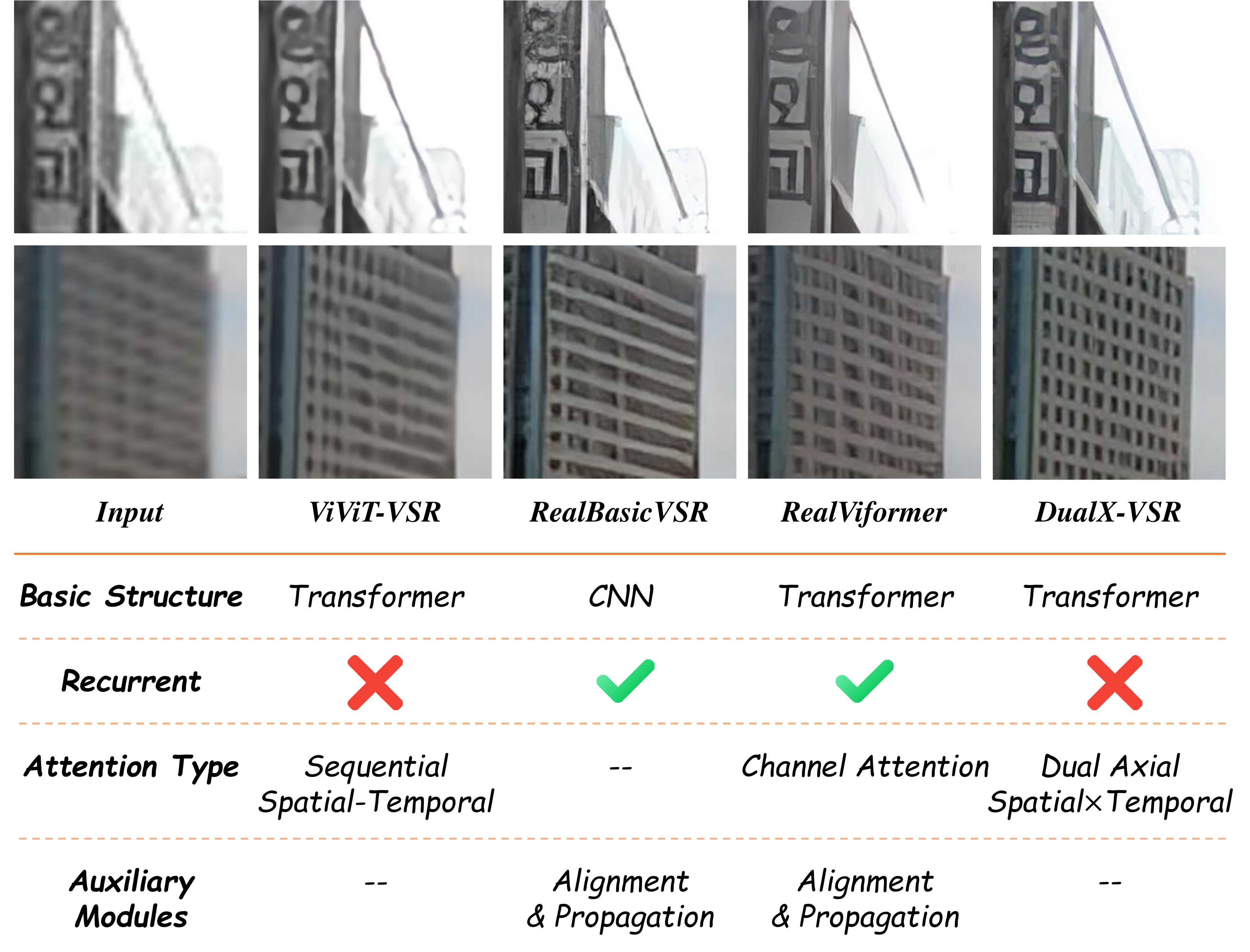}
   \caption{Comparison of different VSR models. ViViT-VSR follows ViViT’s attention design without auxiliary modules but performs poorly due to heavy information compression. RealBasicVSR and RealViformer enhance performance with alignment, propagation modules and recurrent structures, while they face limitations in receptive field and scalability. DualX-VSR retains ViViT’s simplicity and surpasses current models through dual axial spatial$\times$temporal attention.}
   \label{fig:first}
   \vspace{-10pt}
\end{figure}


Real-world VSR aims to restore high-resolution results from low-quality, degraded video inputs that often contain noise, blur, and compression artifacts. So, can directly applying ViViT's architecture to the real-world VSR task yield promising results? As shown in Fig.~\ref{fig:first}, we build a video super-resolution network based on the ViViT structure, namely ViViT-VSR. ViViT-VSR inherits the same video vision transformer framework and sequential spatial-temporal attention structure as ViViT, without incorporating any auxiliary modules (such as optical flow-based alignment). Unfortunately, even with significantly more parameters, ViViT-VSR still performs poorly, exhibiting substantial loss of fine texture details. This limitation arises from the intrinsic characteristics of low-level vision tasks. Unlike high-level tasks, low-level vision tasks demand pixel-level precision and preservation of fine-grained details, which makes them more vulnerable to information loss. Processes such as compressed patchification, tokenization, and sequential spatial-temporal attention in ViViT can result in substantial detail loss, impairing the model’s ability to produce high-quality reconstructions.

To mitigate these issues, ViT-based image and video super-resolution models have integrated specific adjustments, such as smaller patch sizes, CNN-based pixel refinement~\cite{ipt}, and local attention (e.g., Swin Transformer~\cite{swin_transformer}) alongside flow-based feature alignment and fusion modules~\cite{vrt, rvrt, psrt, iart}. With the use of recurrent structure design and auxiliary modules, both CNN-based RealBasicVSR and Transformer-based RealViformer achieve significantly better results than ViViT-VSR, highlighting the effectiveness of these modifications for real-world VSR tasks. However, they still face considerable limitations: local attention restricts receptive field and scalability, while reliance on optical flow for alignment often results in cumulative errors over long sequences or under real-world degradation conditions~\cite{realbasicvsr, videogigagan}.

Encouraged by the recent success of models like Sora~\cite{sora} and VDT~\cite{vdt}, we revisit the video transformer architecture with a focus on real-world VSR. This motivates us to design a straightforward yet effective architecture that bypasses the limitations of optical flow-based alignment and local attention. By examining the core spatial and temporal attention mechanisms in video transformers, we find that existing spatial-temporal attention mechanisms often handle space and time independently and sequentially. This sequential approach fails to fully integrate spatiotemporal information, which limits cohesive video representation. 


In response, we propose a novel Dual Axial Spatial$\times$Temporal Transformer for Real-World Video Super-Resolution (\textbf{DualX-VSR}). Unlike conventional spatial-temporal attention, our approach applies vertical-temporal and horizontal-temporal attention, projecting spatiotemporal information along orthogonal directions to achieve integrated modeling of both spatial and temporal information. This avoids the sequential stacking of spatial and temporal modules, providing a more unified and cohesive representation.


We validate the effectiveness of this approach on real-world VSR tasks. As shown in Fig.~\ref{fig:first}, DualX-VSR retains the simplicity of the ViViT's architecture without additional auxiliary modules and achieves superior performance compared to current VSR methods. By eliminating explicit motion estimation and long-range feature propagation, DualX-VSR provides a robust solution to the unique challenges of real-world video super-resolution, setting a new standard for high-quality VSR in real-world scenarios.

\section{Related Work}
\label{sec:related_work}

\noindent\textbf{Video Super-Resolution (VSR).} Recent advancements in video super-resolution (VSR) predominantly utilize frameworks based on optical flow alignment and propagation. Early approaches rely heavily on optical flow-based warping to align neighboring frames~\cite{spatio_temporal_transformer,Xue_2019}. However, these methods are highly sensitive to inaccuracies in flow estimation, often resulting in significant performance degradation. To mitigate this issue, later methods introduce feature map alignment techniques~\cite{basicvsr} and incorporate optical flow guidance within deformable convolutions~\cite{basicvsr++, edvr}. From VSRT~\cite{vsrt} and VRT~\cite{vrt}, transformer frameworks are introduced to the VSR task, yet they retain the same structure as CNN-based models. Subsequently, many works leverage the unique properties of transformers to optimize this framework. RVRT~\cite{rvrt} introduces deformable attention mechanisms, while PSRT~\cite{psrt} incorporates patch alignment based on the transformer’s inherent alignment capabilities, reducing reliance on precise flow estimation. Additionally, IART~\cite{iart} proposes an implicit resampling-based alignment using coordinate networks and cross-attention, enhancing the preservation of high-frequency details and minimizing spatial distortions.

\noindent\textbf{Real-world VSR.} When degradation shifts from bicubic downsampling to real-world conditions, the task becomes significantly more challenging. Early works such as RealBasicVSR~\cite{realbasicvsr} address these complexities in real-world VSR by introducing an image pre-cleaning stage to improve detail synthesis and reduce artifacts caused by diverse degradations. Recent works also adopt attention-based structures. For example, Upscale-A-Video~\cite{upscaleavideo} enhances local coherence by integrating temporal layers and adds a flow-guided propagation module for global stability.  RealViformer~\cite{realviformer} improves performance by using channel attention to reduce artifact sensitivity and enhance reconstruction quality.

\noindent\textbf{Vision Transformer (ViT) in Video Tasks.} The Vision Transformer (ViT)~\cite{vit} initially revolutionizes 2D image tasks with its transformer architecture, and subsequent ViT-based methods deliver remarkable image processing results. In the nascent exploration of extending Vision Transformers (ViT) to video tasks, ViViT~\cite{vivit} pioneers the use of sequential spatial-temporal attention in video classification, proposing efficient model variants to address the challenges of long video sequences. For other video tasks, Latte~\cite{latte}, VDT~\cite{vdt}, and Sora~\cite{sora} introduce a Latent Diffusion Transformer to generate high-fidelity videos using a transformer architecture over spatiotemporal patches. LaVie~\cite{lavie} capitalizes on a pre-trained ViT to build a text-to-video generation framework, effectively capturing temporal dynamics with temporal self-attention and rotary positional encoding. Additionally, previous work MedT~\cite{medical_transformer} introduces a medical image segmentation model using axial-attention within transformers, which processes height and width separately for efficiency and captures long-range dependencies. These works demonstrate the potential of the ViT architecture while also emphasizing the critical role of attention design in achieving effective performance~\cite{vivit, medical_transformer, latte}.

\section{Investigating Attention Mechanisms in Video Super-Resolution}
\label{sec:attention}


\begin{figure}[t]
  \centering
   \includegraphics[width=0.99\linewidth]{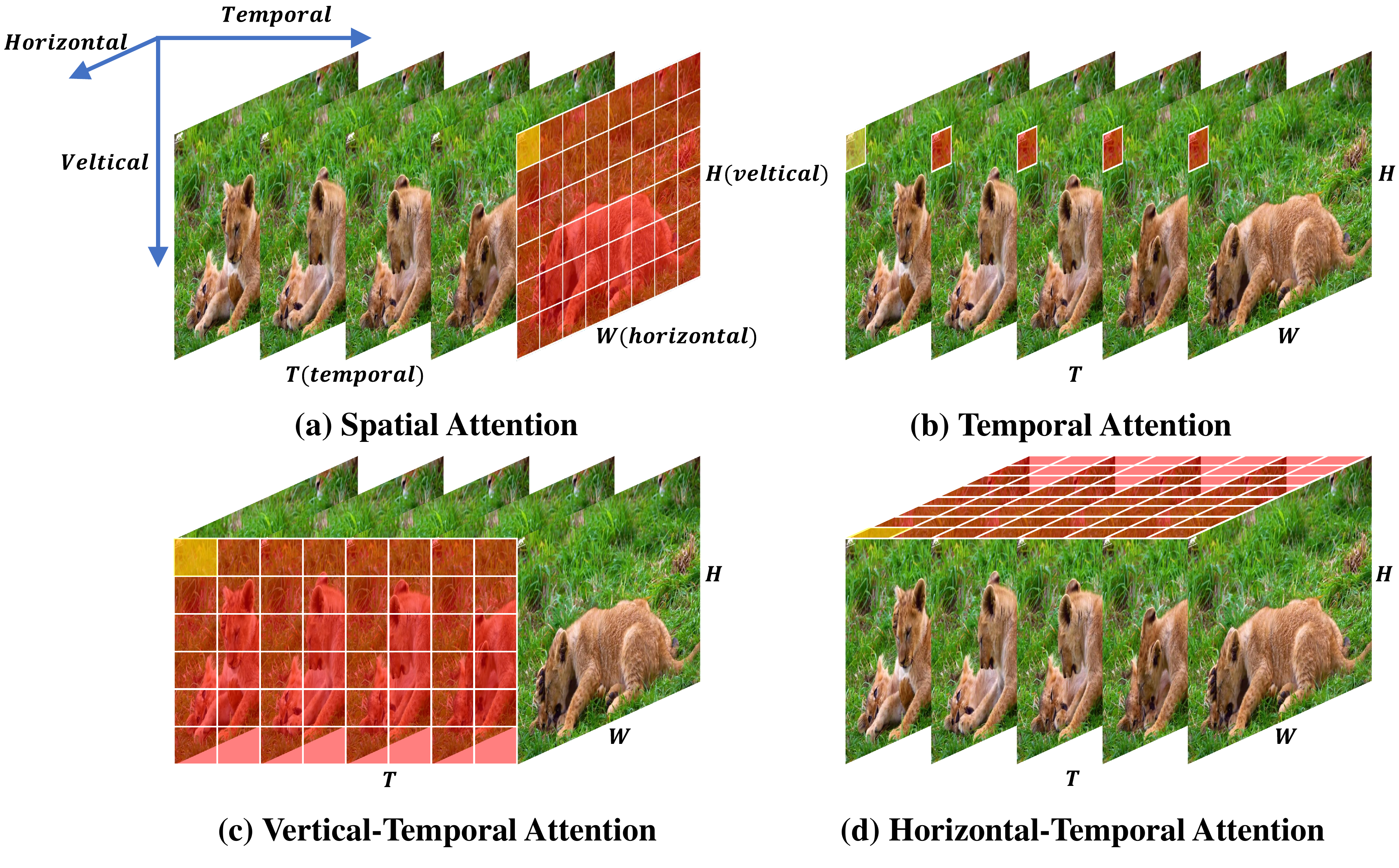}
   \caption{Diagram of different attention. Time (T), Width (W), and Height (H) correspond to Temporal, Horizontal, and Vertical dimensions. \textcolor{orange}{Orange} patch indicates the token being attentioned, while \textcolor{red}{red} patches show its attention range. Spatial and temporal attention lack either temporal or spatial context, while vertical-temporal and horizontal-temporal attention model both dimensions simultaneously, enabling natural spatiotemporal fusion for performance improvement.}
   \label{fig:attention}
   \vspace{-10pt}
\end{figure}

Attention mechanisms are instrumental in video transformer architectures, enabling models to capture both spatial and temporal information essential for video tasks. In this section, we explore various attention mechanisms and analyze their impact on VSR performance.

\subsection{Revisiting Attention Mechanisms in Video Vision Transformers}
\noindent\textbf{Spatial Attention.} The Vision Transformer (ViT)~\cite{vit} introduces a pioneering approach to spatial attention, focusing on capturing spatial dependencies within individual frames.  As shown in Fig.~\ref{fig:attention} (a), ViT divides each input image into non-overlapping patches and applies self-attention across these patches to emphasize static spatial information. However, in video tasks, spatial attention alone has limited utility as it does not account for temporal dependencies between consecutive frames.

\noindent\textbf{Temporal Attention.} To capture essential inter-frame relationships, ViViT~\cite{vivit} incorporates temporal attention specifically for video understanding. As shown in Fig.~\ref{fig:attention} (b), temporal attention computes self-attention across all frames by linking tokens that correspond to the same spatial location across time. This mechanism enables the model to capture motion patterns and dynamic temporal information. However, focusing solely on temporal dependencies can lead to a loss of fine-grained spatial details within frames. Consequently, pure temporal attention structures are rarely used.

\noindent\textbf{Spatial-Temporal Attention.} Spatial-temporal attention aims to combine both spatial and temporal information, enabling the model to focus on critical regions within frames while simultaneously understanding inter-frame dynamics. Many video transformer models apply spatial and temporal attention sequentially.~\cite{vivit, timesformer, latte} For instance, ViViT~\cite{vivit} implements spatial attention layers first, followed by temporal attention layers, while TimeSformer~\cite{timesformer} proposes a divided Space-Time attention mechanism that alternates temporal and spatial attention within each layer. In video generation models like VideoLDM~\cite{videoldm} and Latte~\cite{latte}, spatial and temporal attention blocks are applied in an alternating pattern to improve both spatial quality and temporal consistency. However, these sequential methods still process space and time information separately, which may limit their ability to capture complex, dynamic spatiotemporal patterns.

\noindent\textbf{Local Window-Based Attention.} Local window-based attention is widely used in video super-resolution models like RVRT~\cite{rvrt}, PSRT~\cite{psrt} and IART~\cite{iart}. Unlike high-level tasks such as video classification and understanding, high-frequency texture details are crucial in low-level tasks. Directly applying the same plain vision transformer used in high-level tasks can lead to a significant performance drop. Consequently, starting from Video Restoration Transformer (VRT)~\cite{vrt}, a series of methods have widely adopted the local window-based attention mechanism. Specifically, this attention model utilizes the Swin Transformer architecture~\cite{swin_transformer} and computes attention on video frame clips within a local window. Thanks to this refined structural design, the model demonstrates improved performance on low-level tasks. However, this attention mechanism suffers from limited receptive field, poor scalability and challenges in extending to larger contexts, restricting its ability to capture long-range dependencies effectively.

\subsection{Effects of Different Attention Mechanisms on VSR Performance}

Regarding the great performance of plain vision transformer architectures in other video tasks, we extend this approach to Video Super-Resolution and conduct a detailed investigation into the effects of different attention mechanisms on VSR performance. To ensure that other factors are excluded, we choose the simplest and most extensible video vision transformer architecture ViViT, and adopt it into a video super-resolution model ViViT-VSR. We conducted experiments by training models on REDS dataset~\cite{reds} and testing them on REDS4 dataset~\cite{reds}, varying only the type of attention mechanism. Results are presented in Tab.\ref{tab:attention_comparison}.



\noindent\textbf{Look at the Space or Time: (Only) Spatial or Temporal Attention.} Adopting only spatial or temporal attention limits the model to focusing on a single dimension, either space or time, leading to overall poor performance as shown in Tab.~\ref{tab:attention_comparison}. Interestingly, temporal attention outperforms spatial attention in this case. This is because the REDS4 dataset lacks repetitive structural patterns within each frame, resulting in suboptimal performance for spatial attention. Temporal attention, on the other hand, benefits from two factors: first, some spatial texture information is inherently injected during the patchification and tokenization process; second, it captures tokens at the same spatial locations across different time steps, making it easier to learn patterns that have structural and textural similarities, leading to better results.


\begin{table}[t]
\centering
\resizebox{0.48\textwidth}{!}{
\begin{tabular}{c | c c c c}
\toprule[1.5pt] 
NO. & Attention Type & Runtime (ms) & PSNR$\uparrow$ & SSIM$\uparrow$  \\ \hline
1 & Only Spatial  & 55 & 27.96 & 0.7960\\ 
2 & Only Temporal   & 45 & 28.17 & 0.8041\\ 
3 & Spatial-Temporal   & 50 & 28.22 & 0.8063\\ 
4 & Vertical-Temporal   & 45 & 28.22 & 0.8059\\ 
5 & Horizontal-Temporal    & 45 & 28.32 & 0.8072\\  
\rowcolor{gray!15}6 & \textbf{Dual Axial Spatial$\times$Temporal}  & 45 & \textbf{28.38} & \textbf{0.8099} \\ \toprule[1.5pt]
\end{tabular}
}
\caption{Effects of different attention mechanisms on VSR tasks. We compute runtime per frame on 320 $\times$ 180 to 1280 $\times$ 720 on REDS~\cite{reds} dataset. All models use the same attention module implementation with 24 total units, evenly distributed for multiple attention types. Within the same parameter size, dual axial spatial$\times$temporal attention demonstrates a significant improvement both in performance and runtime.}
\label{tab:attention_comparison}
\end{table}

\noindent\textbf{Look at the Space then Time: (Sequential) Spatial-Temporal Attention.} By integrating spatial and temporal attention in a sequential manner, spatial-temporal attention achieves performance improvement in Tab.~\ref{tab:attention_comparison}. This phenomenon is consistent with results in high-level video tasks, highlighting that for VSR, it is crucial for a video vision transformer to effectively capture both temporal and spatial information simultaneously.


\begin{figure}[htbp]
  \centering
  \vspace{-10pt}
   \includegraphics[width=0.99\linewidth]{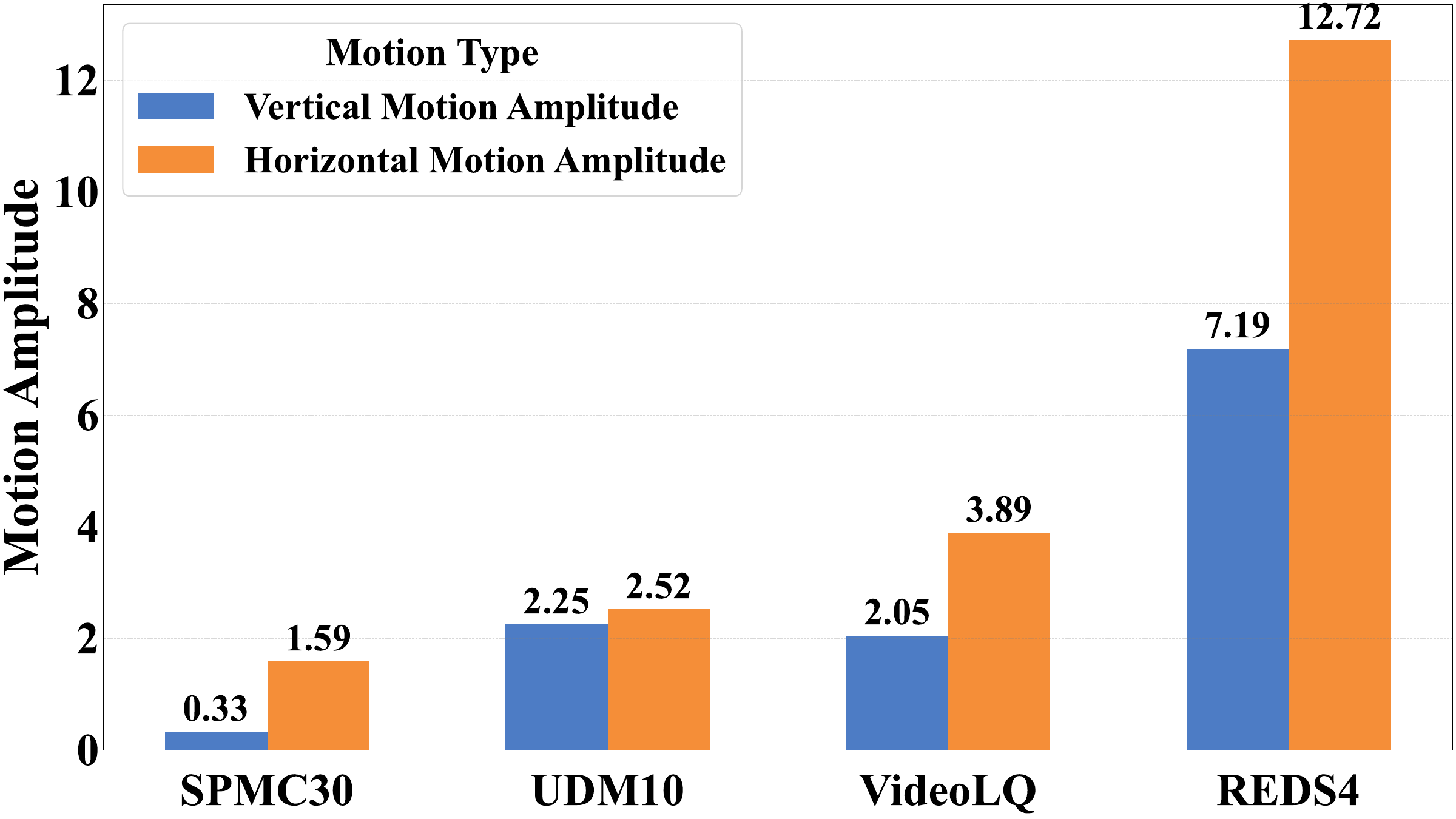}
   \caption{Average motion amplitude across different video datasets. We estimate optical flow and calculate average horizontal and vertical motion amplitudes across datasets with FlowFormer~\cite{flowformer}. While motion amplitudes vary, average amplitude of horizontal motion consistently exceeds vertical motion, especially in SPMC30 and REDS4. This suggests that models optimized for horizontal motion tend to outperform across most video datasets.}
   \label{fig:motion_amplitude}
\end{figure}

\noindent\textbf{Look at the Space and Time Simultaneously: Dual Axial Spatial$\times$Temporal Attention.} Even though spatial-temporal attention has shown improved results~\cite{vivit, timesformer, latte}, it fundamentally processes spatial and temporal information independently before fusion. Is there a more suitable approach for video tasks that directly integrates temporal and spatial information? The attention mechanism used in Vidu ~\cite{vidu} is a successful example of spatial-temporal integration, which we refer as omnidirectional spatial-temporal attention. Vidu directly combines information from the Height-Width-Time dimensions into 3D video patches and tokens, with each token containing both spatial and temporal information. Omnidirectional spatial-temporal attention directly operates on these 3D video tokens, enabling the model to capture both spatial and temporal dependencies simultaneously, thus enhancing the ability to model dynamic patterns in video sequences.


Unfortunately, directly applying omnidirectional spatial-temporal attention to video super-resolution presents an unmanageable computational cost. Therefore, we seek to decompose the original three dimensions into two separate components along the Height-Time and Width-Time dimensions, resulting in a more efficient attention mechanism—Dual Axial Spatial$\times$Temporal Attention—that reduces computational complexity while maintaining the ability to capture critical spatial-temporal dependencies.

Dual Axial Spatial$\times$Temporal Attention integrates Vertical-Temporal attention and Horizontal-Temporal attention in a sequential manner, as shown in Fig.~\ref{fig:attention} (c) and Fig.~\ref{fig:attention} (d), respectively. Specifically, vertical-temporal attention performs tokenization and attention computation along the Height-Time dimensions, capturing vertical motion and spatial textures, while horizontal-temporal attention performs along the Width-Time dimensions, capturing horizontal motion and spatial textures. As shown in Tab.~\ref{tab:attention_comparison}, both vertical-temporal attention and horizontal-temporal attention demonstrate comparable or even superior performance compared to sequential spatial-temporal attention. Horizontal-temporal attention shows a large improvement over vertical-temporal attention, as the majority of motion in the test videos occurs more prominently in the horizontal direction than in the vertical direction. As shown in Fig.~\ref{fig:motion_amplitude}, we compute optical flow for four datasets using FlowFormer~\cite{flowformer} and analyze the motion amplitudes in the horizontal and vertical directions. The results reveal that the average horizontal motion amplitude across all datasets is substantially greater than the vertical motion amplitude. Consequently, horizontal-temporal attention, which focuses on motion in the horizontal direction, achieves better performance. This phenomenon further validates the effectiveness and physical relevance of the proposed dual axial attention. By combining these two attention mechanisms, dual axial spatial$\times$temporal attention achieves significant performance improvements over sequential spatial-temporal attention~\cite{vivit}. Furthermore, since the number of tokens along the temporal dimension is generally much smaller than that along the spatial dimensions, dual axial spatial$\times$temporal attention also greatly enhances computational efficiency, resulting in a shorter runtime. 




\section{Methodology}
\label{sec:method}

\begin{figure*}[t]
  \centering
   \includegraphics[width=0.99\linewidth]{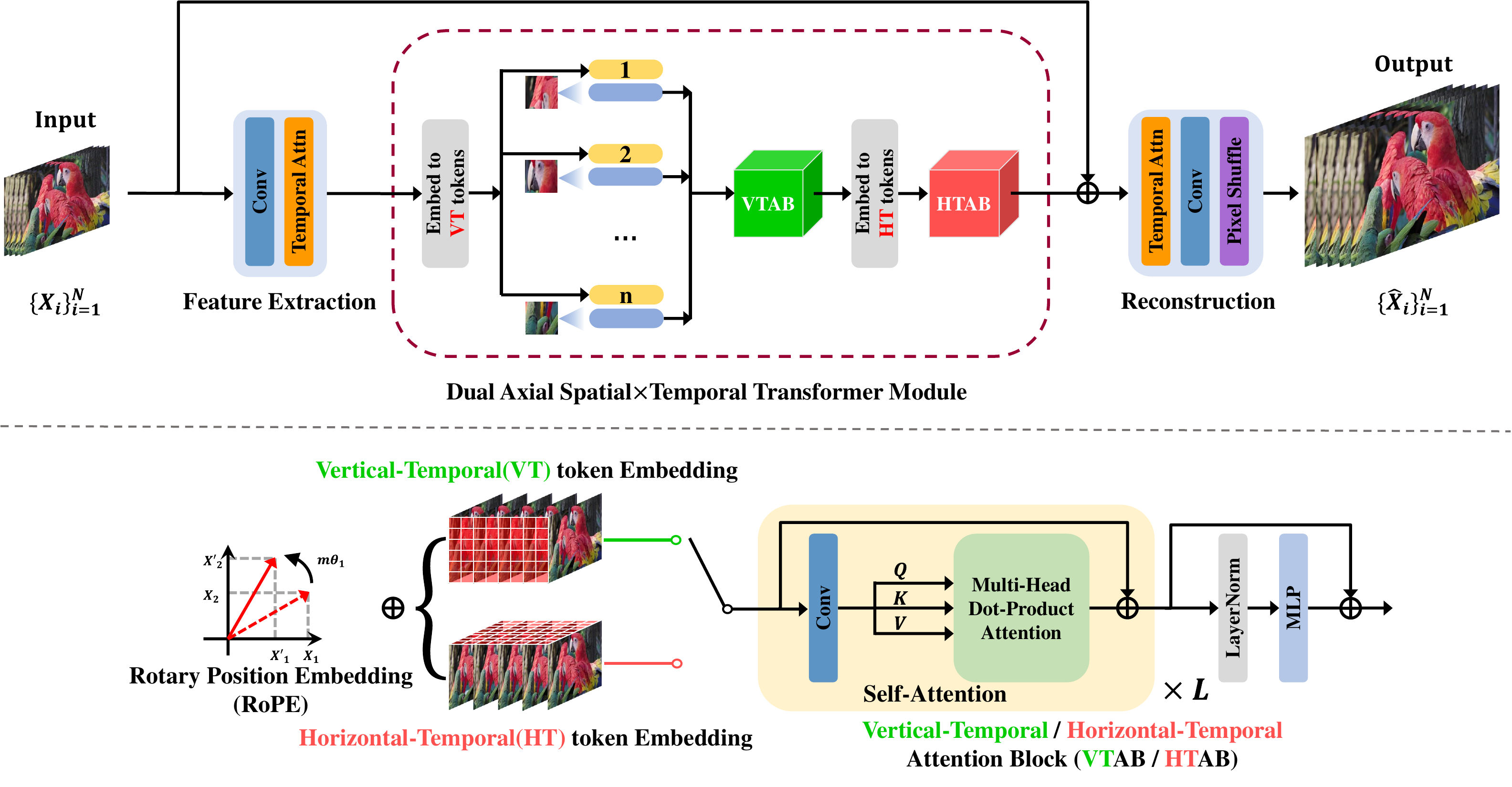}
   \caption{Overview of the proposed DualX-VSR. The top section shows the pipeline, which converts low-quality video to high-quality video through the Input Video Embedding, Dual Axial Spatial$\times$Temporal Transformer Module and Spatiotemporal Reconstruction Module. The bottom section details the Dual Axial Spatial$\times$Temporal Transformer Module, comprising the Vertical-Temporal Attention Block (VTAB) and Horizontal-Temporal Attention Block (HTAB). The left branch indicates different dimensional transformations during token embedding, while the right shows that VTAB and HTAB share the same internal self-attention structure.}
   \label{fig:pipeline}
\end{figure*}


\subsection{Overview}

The overall pipeline of Dual Axial Spatial$\times$Temporal Transformer is shown in Fig.~\ref{fig:pipeline}. Let's denote the input video as $\{X_i\}_{i=1}^N \in \mathbb{R}^{B \times C \times N \times H \times W}$, where $i$ represents the frame index, $B$ is the batch size, $C$ is the number of channels, $N$ is the total number of frames, and $H, W$ are the height and width of each frame. After processing through the network, we obtain the output high-quality video $\{\hat{X}_i\}_{i=1}^N \in \mathbb{R}^{B \times C \times N \times sH \times sW}$, where $s$ denotes the upscale factor, typically set to 4. We employ a ViViT-based architecture~\cite{vivit}, incorporating input video embedding and spatiotemporal reconstruction module to better suit the characteristics of the VSR task. For transformer module we introduce dual axial spatial$\times$temporal attention to achieve enhanced fusion and modeling of spatiotemporal information. Additionally, we employ a pretrain-finetune training strategy to adapt the model to real-world degradations.

\subsection{Input Video Embedding}
\noindent\textbf{Spatiotemporal preprocessing.} As shown in Fig.~\ref{fig:pipeline}, for the input video $\{X_i\}_{i=1}^N \in \mathbb{R}^{B \times C \times N \times H \times W}$ , we first use a 2D convolution to capture the spatial characteristics of each frame. Specifically, we obtain shallow features $\{F_i\}_{i=1}^N \in \mathbb{R}^{B \times d \times N \times H \times W}$, where $d$ represents the dimension of the shallow features. Next, we perform a $1\times1$ patchification of these features, followed by temporal attention to enhance the temporal continuity. 

\noindent\textbf{Token embedding.} After spatiotemporal preprocessing, we segment the  feature sequence into patches $\{F'_i\}_{i=1}^N \in \mathbb{R}^{B \times d \times n_N \times n_H \times n_w}$, where $n_N = \left\lfloor \frac{N}{n} \right\rfloor, n_H = \left\lfloor \frac{H}{h} \right\rfloor, n_W = \left\lfloor \frac{W}{w} \right\rfloor$,  $n$, $h$ and $w$ represent the patch sizes along the time, height and width dimensions, respectively. Then we employ Rotary Positional Encoding (RoPE)~\cite{rope} to embed the patches into tokens $\{Z_i\}_{i=1}^N \in \mathbb{R}^{B \times D \times n_N \times n_H \times n_w} $, where $D$ represents for the embedding dimension of attention block. Specifically, given the $u^{th}$ key and $v^{th}$ query vector as $\textbf{q}_u, \textbf{k}_v \in \mathbb{R}^{|D|}$, RoPE multiplies the bias to the key or query vector in the complex vector space:

\begin{equation}
f_q(\textbf{q}_u, u) = e^{iu \Theta} \textbf{q}_u, \quad f_k(\textbf{k}_v, v) = e^{iv \Theta} \textbf{k}_v,
\end{equation}

where $\Theta = \text{Diag}(\theta_1, \cdots , \theta_{|D|/2})$ is the rotary frequency matrix with $\theta_d = b^{-2d/|D|}$ and rotary base $b = 10000$. The attention score is calculated as:

\begin{equation}
A_v = \operatorname{Re} \langle f_q(\textbf{q}_u, u), f_k(\textbf{k}_v, v) \rangle.
\end{equation}

RoPE offers several advantages, such as better handling of long-range dependencies and maintaining spatial and temporal coherence, which aligns well with the requirements of VSR tasks.

\subsection{Dual Axial Spatial$\times$Temporal Attention}
\label{dual_axial_spatial_x_temporal_attention}

Dual axial spatial$\times$temporal attention consists of two types of attention: vertical-temporal attention and horizontal-temporal attention. As shown in Fig.~\ref{fig:pipeline}, in the Dual Axial Spatial × Temporal Transformer module, we first perform token embedding along the height-time dimension, rearranging token dimension from $[B \, D \, n_N \, n_H \, n_W]$ to $[(B \, D \, n_W) \, n_H \, n_N]$. The token is then fed into the Vertical-Temporal Attention Block (VTAB) to compute self-attention in the vertical-temporal plane, effectively modeling spatial textures and vertical motion in the video. Subsequently, token embedding is conducted along the width-time dimension, rearranging token dimension from $[(B \, D \, n_W) \, n_H \, n_N]$ to $[(B \, D \, n_H) \, n_W \, n_N]$. Then the Horizontal-Temporal Attention Block (HTAB) is applied to capture spatial textures and horizontal motion in the video. This enables the effective utilization of spatial features and motion characteristics in both vertical and horizontal directions, enhancing the model's ability to capture intricate details and temporal dynamics. 


\subsection{Spatiotemporal Reconstruction Module}
In conventional VSR models, the reconstruction module typically processes frames individually, lacking the integration of temporal information~\cite{vrt, psrt, iart, realviformer}. We modify this module to enable simultaneous fusion of spatial and temporal information during reconstruction. As shown in Fig.~\ref{fig:pipeline}, after decoding and unpatchifying the tokens output by the dual axial spatial$\times$temporal transformer module, we first apply temporal attention to integrate temporal information. Then, frame-by-frame reconstruction is carried out using operations such as 2D convolution and pixel shuffle. Experiments show that 3D convolution does not improve performance over 2D convolution and significantly increases computational cost, so we use 2D convolution for reconstruction. This approach allows the model to incorporate both spatial and temporal information, resulting in video outputs $\{\hat{X}_i\}_{i=1}^N \in \mathbb{R}^{B \times C \times N \times sH \times sW}$ with enhanced spatiotemporal quality.

\subsection{Training Strategy}
Due to the complexity of real-world degradations, training directly from scratch often fails to achieve optimal results. Many studies~\cite{realesrgan, hat, sequential} have highlighted that starting with simple degradations and progressively transitioning to more complex ones can help the model perform better on real-world degradations. Therefore, we adopt a pretrain-finetune strategy on DualX-VSR. Specifically, we first pre-train the model on a dataset synthesized with only bicubic downsampling, and then fine-tune the model on the same dataset with more complex degradations. Experimental results show that this training strategy significantly improves model performance, as detailed in Sec.~\ref{sec:ablation}.

\begin{table*}[htbp]
\centering
\resizebox{0.99\textwidth}{!}{
\begin{tabular}{c | ccc | ccc | ccc | cc}
\toprule[1.5pt] 
\multirow{2}{*}{Model} & \multicolumn{3}{c}{REDS4\cite{reds}} & \multicolumn{3}{c}{UDM10\cite{udm}} & \multicolumn{3}{c}{SPMC30\cite{spmc}} &  \multicolumn{2}{c}{VideoLQ\cite{realbasicvsr}} \\
\cmidrule(r){2-4} \cmidrule(r){5-7} \cmidrule(r){8-10} \cmidrule(r){11-12} 
& PSNR$\uparrow$ & SSIM$\uparrow$ & LPIPS$\downarrow$ 
& PSNR$\uparrow$ & SSIM$\uparrow$ & LPIPS$\downarrow$ 
& PSNR$\uparrow$ & SSIM$\uparrow$ & LPIPS$\downarrow$ 
& LIQE$\uparrow$ & NIQE$\downarrow$ \\
\midrule
Real-ESRGAN~\cite{realesrgan} & 22.3927 & 0.5984 & 0.3095 & 26.8230 & 0.8088 & 0.2328 & 22.6363 & 0.6317 & 0.2987 & 2.4084 & 4.2043 \\
SD$\times$4 Upscaler~\cite{sdupsampler} & 21.3590 & 0.5181 & 0.3484 & 24.1116 & 0.6969 & 0.2798 & 20.5567 & 0.4790 & 0.3573 & 1.6776 & 5.8093 \\
DiffBIR~\cite{diffbir} & 23.2778 & 0.5898 & 0.3251 & 26.9274 & 0.7681 & 0.2881 & 22.8967 & 0.5815 & 0.3252 & 2.0023 & 4.9699 \\ \hline
ViViT-VSR (MSE)~\cite{vivit} & 25.0982 & 0.6839 & 0.4512 & 29.3142 & 0.8499 & 0.2537 & 24.9858 & 0.7105 & 0.4103 & 1.3301 & 7.1305 \\
\rowcolor{gray!15}\textbf{DualX-VSR (MSE)} & \textbf{26.0300} & \textbf{0.7092} & 0.3896 & \textbf{30.1271} & \textbf{0.8701} & 0.2225 & \textbf{25.8085} & \textbf{0.7416} & 0.3481 & 1.8052 & 6.7903 \\ \hline
RealVSR~\cite{realvsr} & 21.1990 & 0.5954 & 0.6632 & 23.3811 & 0.7669 & 0.5109 & 21.5877 & 0.6208 & 0.6102 & 1.0479 & 5.9744 \\
DBVSR~\cite{dbvsr} & 22.3794 & 0.5884 & 0.4996 & 25.6422 & 0.7657 & 0.3087 & 21.9512 & 0.5774 & 0.4192 & 1.1144 & 6.2782 \\
RealBasicVSR~\cite{realbasicvsr} & 23.9997 & \textcolor{blue}{0.6505} & 0.2551 & 27.6895 & 0.8180 & 0.2337 & \textcolor{blue}{24.0215} & \textcolor{blue}{0.6678} & \textcolor{red}{0.2691} & \textcolor{red}{3.0239} & \textcolor{red}{3.6921} \\ 
RealViformer~\cite{realviformer} & \textcolor{blue}{24.0925} & 0.6446 & \textcolor{blue}{0.2473} & \textcolor{blue}{28.1240} & \textcolor{red}{0.8230} & \textcolor{red}{0.2183} & 23.9038 & 0.6567 & 0.2755 & 2.4022 & 4.0581 \\
Upscale-A-Video~\cite{upscaleavideo} & 23.0456 & 0.5906 & 0.3891 & 27.0560 & 0.7696 & 0.2774 & 21.8967 & 0.5423 & 0.3676 & 1.3982 & 5.1247 \\ 
VEnhancer~\cite{venhancer} & 20.7839 & 0.5565 & 0.4663 & 19.6780 & 0.5200 & 0.4143 & 22.7876 & 0.7214 & 0.3091 & 1.7693 & 5.6064 \\ 
\rowcolor{gray!15}\textbf{DualX-VSR (GAN)} & \textcolor{red}{24.4541} & \textcolor{red}{0.6550} & \textcolor{red}{0.2442} & \textcolor{red}{28.1883} & \textcolor{blue}{0.8181} & \textcolor{blue}{0.2284} & \textcolor{red}{24.3219} & \textcolor{red}{0.6786} & \textcolor{blue}{0.2748} & \textcolor{blue}{2.5515} & \textcolor{blue}{4.0272} \\
\bottomrule[1.5pt] 
\end{tabular}
}
\caption{Quantitative comparison across different benchmarks, \textit{i.e.}, synthetic datasets (REDS4, UDM10, SPMC30) and real-world datasets (VideoLQ). Our appoarches are marked with \colorbox{gray!15}{gray} background. The best and second performances with adversarial loss are marked in \textcolor{red}{red} and \textcolor{blue}{blue}, respectively. The best performance achieved by the MSE models is highlighted in \textbf{bold}. Our DualX-VSR (MSE) and DualX-VSR (GAN) demonstrate highly competitive performance across synthetic and real-world datasets.}
\label{tab:comparison_with_sota}
\end{table*}

\section{Experiments}
\label{sec:Experiments}

\subsection{Datasets and Implementation}
\noindent\textbf{Datasets.} All of our training is conducted on the REDS dataset~\cite{reds}. For degradation synthesis on training datasets, we generated LQ-HQ video pairs following the pipeline of RealBasicVSR~\cite{realbasicvsr}. For synthetic test datasets, we selected REDS4~\cite{reds}, UDM10~\cite{udm} and SPMC30~\cite{spmc}, using the same degradation settings as in training for evaluation. For real-world datasets, we use VideoLQ~\cite{realbasicvsr}.

\noindent\textbf{Training Details.} Referring to previous studies~\cite{realesrgan, hat, sequential}, we adopt a progressive training strategy. Our training process consists of three stages. The first stage is conducted on 8 NVIDIA A100-80G GPUs with a batch size of 32 for 500k iterations. The input sequence length is set to 16 frames, and the training data is cropped to a size of $64\times64$. The learning rate is set to $5\times 10^{-5}$, using the AdamW~\cite{adam} optimizer. We then fine-tune the network on the REDS dataset with degradation for 150k iterations, reducing the batch size to 4, with 16 input frames and a crop size of $112\times112$, while keeping the learning rate and optimizer unchanged. This stage produces the DualX (MSE) model. In the third stage, we continue fine-tuning on the same dataset in previous stage with perceptual loss and adversarial loss for additional 25k iterations. The batch size remains 4 and the number of input frames is reduced to 8, and the crop size is set to $64\times64$. The learning rate for both the generator and discriminator is set to $1\times 10^{-5}$.

\noindent\textbf{Evaluation Metrics.} We employ various metrics to evaluate the quality of the video super-resolution results. Following previous works, for synthetic datasets with LQ-HQ pairs, we empoly PSNR, SSIM and LPIPS~\cite{psnr}. For real-world test datasets, we calculate commonly used non-reference metrics, \textit{i.e.}, NIQE~\cite{niqe} and LIQE~\cite{liqe}. For temporal consistency, we employ $E_{warp}$~\cite{ewarp} and BackgroundConsistency (from VBench~\cite{vbench,vbench++}).

\begin{figure*}[t]
  \centering
   \includegraphics[width=0.99\linewidth]{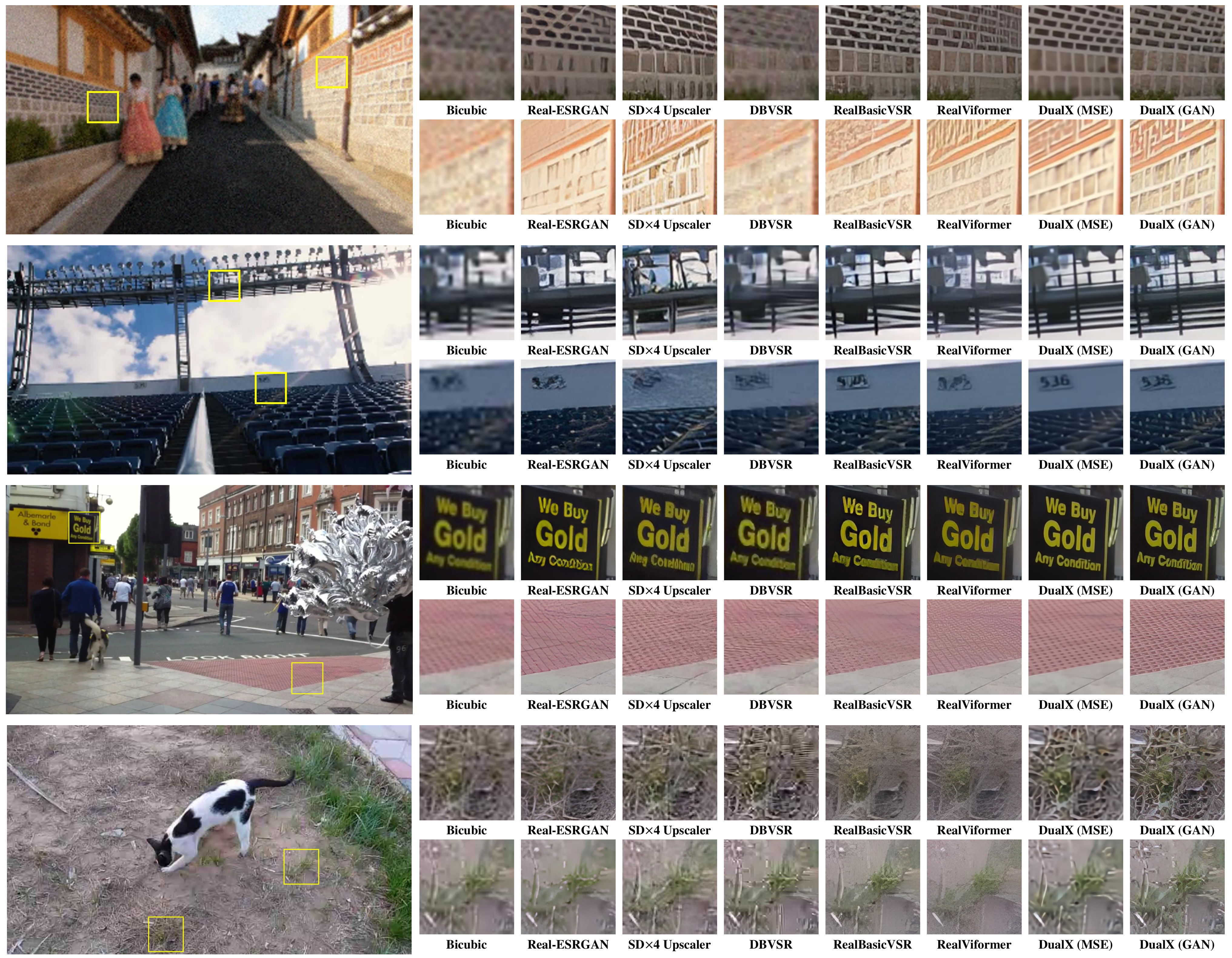}
   \caption{Qualitative comparisons on both synthetic datasets and real-world datasets. The top two comparisons are from the REDS and UDM10, while the bottom two are from VideoLQ. Our DualX-VSR (MSE) and DualX-VSR (GAN) demonstrate excellent visual performance on both synthetic and real-world datasets, effectively removing degradations and achieving high-fidelity results.}
   \label{fig:visual_comparison}
\end{figure*}

\subsection{Comparisons with State-Of-The-Art Methods}
To verify the quality of our DualX-VSR model, we compare it with existing state-of-the-art image and video super-resolution models, including Real-ESRGAN~\cite{realesrgan}, SD$\times$4 Upscaler~\cite{sdupsampler}, DiffBIR~\cite{diffbir}, RealVSR~\cite{realvsr}, DBVSR~\cite{dbvsr}, RealBasicVSR~\cite{realbasicvsr}, RealViformer~\cite{realviformer}, Upscale-A-Video~\cite{upscaleavideo} and VEnhancer~\cite{venhancer}. Additionally, we train a video super-resolution model, ViViT-VSR, directly adopted from ViViT~\cite{vivit} and configured identically to DualX-VSR.

\noindent\textbf{Quantitative Evaluation.} As shown in Tab.~\ref{tab:comparison_with_sota}, DualX-VSR (MSE) achieves the highest PSNR and SSIM across all synthetic datasets, significantly outperforming other methods. Furthermore, compared to ViViT-VSR, DualX-VSR (MSE) shows substantial improvements across all datasets and metrics, highlighting the effectiveness of our architecture design and training strategy. For DualX-VSR (GAN), it achieves the highest PSNR against the models with generative capabilities on synthetic datasets and attains the highest SSIM and lowest LPIPS on REDS4 and SPMC30, indicating that our model delivers results with both high fidelity and perceptual quality. Additionally, our model shows competitive performance on real-world dataset VideoLQ. As shown in Tab.~\ref{tab:complexity_temporal}, DualX-VSR also demonstrates competitive performance in temporal consistency. Notably, traditional methods exhibit good consistency, while sacrificing fine-grained details.  The comprehensive performance underscores the effectiveness of our approach. 



\noindent\textbf{Qualitative Evaluation.} We present visual results on both synthetic and real-world datasets in Fig.~\ref{fig:visual_comparison}. Compared to existing state-of-the-art methods, our model demonstrate significant abilities in removing degradations and generating realistic details, while maintaining high fidelity. Specifically, DualX-VSR (MSE) excels in achieving the highest fidelity and degradation removal, whereas DualX-VSR (GAN) shows superior detail generation capabilities. Notably, the SD$\times$4 Upscaler~\cite{sdupsampler} lacks degradation removal capabilities, which results in subpar performance in real-world scenarios. For more results, please refer to the Supp.


\noindent\textbf{Complexity \& Scalability.} DualX-VSR explores a pure transformer-based architecture for real-world VSR, which increases the parameter size but gains better performance. DualX-VSR still perform efficiency, such as runtime and memory access costs (MACs), as shown in Tab.~\ref{tab:complexity_temporal}. Benefiting from this structure, DualX-VSR shows appealing scalability compared to traditional models. In contrast, despite scaling up to the same parameter scale, RealBasicVSR and RealViformer exhibit inferior performance, MACs, and runtime compared to DualX-VSR (see Tab.~\ref{tab:scalability}).

\subsection{Exploration on DualX-VSR Structure}
\label{sec:ablation}

\begin{figure}[htbp]
  \centering
   \includegraphics[width=0.99\linewidth]{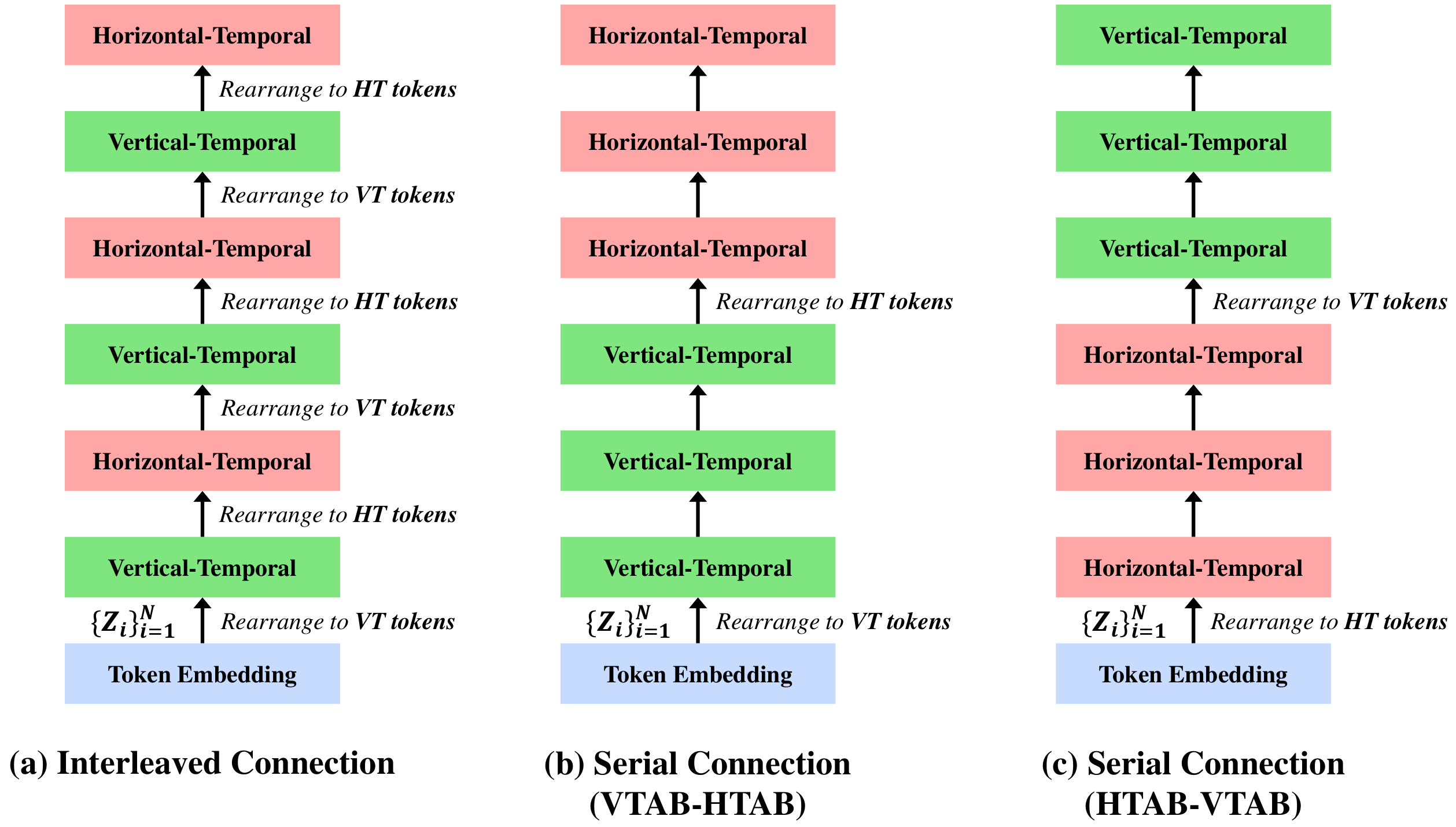}
   \caption{Different structures of attention blocks. (a) illustrates the \textbf{Interleaved Connection}, where the blocks are alternately computed and connected. (b) and (c) depict the \textbf{Serial Connection}, where the blocks are computed sequentially and then combined. }
   \label{fig:connection}
\end{figure}

\noindent\textbf{Attention Block Connection.} For VTAB and HTAB, two connection strategies are typically employed: serial connection and interleaved connection, as illustrated in Fig.~\ref{fig:connection}. As highlighted in generative models such as Latte~\cite{latte}, interleaved connections often yield better performance. However, as shown in Tab.~\ref{tab:ablation_2} (a) and (d), our experiments indicate that serial connection outperforms interleaved connection, whereas DualX-VSR utilizes serial connection. 

\noindent\textbf{Pre-Feature Extraction.} In Input Video Embedding, we employ 2D convolution to extract spatial features from each frame, while temporal continuity is preserved via temporal attention. As shown in Tab.~\ref{tab:ablation_2} (b) and (d), 3D convolution performs slightly worse than 2D convolution and requires more computational resources. Consequently, DualX-VSR adopts 2D convolution for pre-feature extraction.

\noindent\textbf{Attention Block Arrangement.} Arrangement designs of VTAB and HTAB are illustrated in Fig.~\ref{fig:connection} (b) and (c). DualX-VSR incorporates spatiotemporal information within symmetric attention blocks, ensuring that arrangement has minimal impact on performance. Experiments demonstrate that changing arrangement results in a negligible performance drop, as shown in Tab.~\ref{tab:ablation_2} (c) and (d).

\begin{table}[htbp]
    \centering
    \resizebox{0.5\textwidth}{!}{%
    \begin{tabular}{c | c c c | c}
    \toprule[1.5pt] 
         Model & Params(M) & Runtime(s) & MACs(G) &  $E_{warp}^*$\cite{ewarp}$\downarrow$/BC\cite{vbench,vbench++}$\uparrow$ \\ \hline
        
        RealVSR\cite{realvsr} & 2.69 & 0.83 & 90.82 &  1.069 / \textcolor{blue}{\textbf{0.969}}\\ 
        RealBasicVSR\cite{realbasicvsr} & 6.29 & 0.07 & 424.51 &  0.791 / 0.967\\ 
        RealViformer\cite{realviformer} & 5.82 & 0.04 & 160.16 & \textcolor{red}{\textbf{0.571}} / 0.958  \\ 
        Upscale-A-Video\cite{upscaleavideo} & 691.04 & 699.50 & $\backslash$ & 1.347 / 0.963\\ 
        \rowcolor{gray!15}DualX-VSR (GAN) & 127.95 & 0.40 & 99.41 & \textcolor{blue}{\textbf{0.631}} / \textcolor{red}{\textbf{0.971}} \\ 
        \bottomrule[1.5pt] 
    \end{tabular}
    }
    \caption{Comparison of complexity and temporal consistency on SPMC30 dataset. $E_{warp}^*$ denotes $E_{warp}(\times 10^{-3})$. BC refers to BackgroundConsistency in VBench~\cite{vbench,vbench++}.  We compute runtime per frame on 320 $\times$ 180 to 1280 $\times$ 720. MACs are calculated on an input of 16 frames of $64\times64$ image.}
    \label{tab:complexity_temporal}
\end{table}


\begin{table}[htbp]
    \centering
    \resizebox{0.5\textwidth}{!}{%
    \begin{tabular}{c | c c c | c c}
    \toprule[1.5pt] 
         Model & Params (M) & Runtime(s) & MACs(G) & PSNR$\uparrow$  & SSIM$\uparrow$\\ \hline
        RealBasicVSR* (scaling up)\cite{realbasicvsr} & 135.59 & 0.46 & 9195.52 & 25.09 & 0.6761 \\ 
        RealViformer* (scaling up)\cite{realviformer} & 131.32 & 0.44 & 2424.48  & 25.41 & 0.6860  \\ 
        \rowcolor{gray!15}DualX-VSR (MSE)  & 127.95 & 0.40 & 99.41 & \textbf{26.03} & \textbf{0.7092} \\ 
        \bottomrule[1.5pt] 
    \end{tabular}
    }
    \caption{Comparison of scalability. Models with * are scaled up and trained with MSE loss, using the same settings as DualX-VSR.}
    \label{tab:scalability}
\end{table}

\begin{table}[htbp]
\centering
\resizebox{0.5\textwidth}{!}{%
\begin{tabular}{c | c c | c c}
\toprule[1.5pt] 
Exp. & Attention Block Connection(Arrangement) & Pre-Feature Extraction & PSNR $\uparrow$ & SSIM $\uparrow$ \\ \hline
(a) & Interleaved Connection ($\backslash$)  & 2D Conv. & 25.94 & 0.7048 \\ 
(b) & Serial Connection (VTAB-HTAB) & 3D Conv.  & 25.99 & 0.7069 \\ 
(c) & Serial Connection (HTAB-VTAB) & 2D Conv.  & 26.01 & 0.7086 \\
\rowcolor{gray!15}(d) & Serial Connection (VTAB-HTAB) & 2D Conv.  & \textbf{26.03} & \textbf{0.7092} \\
\bottomrule[1.5pt] 
\end{tabular}%
}
\caption{Different designs of DualX-VSR structure. The final design of DualX-VSR (d) yields the best performance.}
\label{tab:ablation_2}
\end{table}

\section{Conclusion}
\label{sec:Conclusion}


ViT-based models' success in video understanding and generation tasks inspire our exploration of transformer-based models for real-world VSR. We introduce DualX-VSR, a novel transformer-based approach that eliminates motion compensation. Based on ViViT's architecture, we employ Dual Axial Spatial$\times$Temporal attention mechanism to integrate spatial and temporal information. This simplified design overcomes the limitations of optical flow-based alignment, propagation and local window-based attention. Experiments demonstrate that DualX-VSR sets a new standard for high-quality real-world VSR. We believe further refinement of transformer architectures for low-level vision tasks could enable more efficient and robust solutions.


\noindent\textbf{Limitations.} Due to limited computational resources, we are unable to fully scale up DualX-VSR. A thorough investigation of its performance at larger scales and under more challenging conditions will be conducted in future work.

%% file: sec/supp.tex
\section{Overview}

In the Appendix, we provide additional details and experiments to further validate our proposed DualX-VSR. Specifically, we provide details and discussion on the architecture and Dual Axial Spatial$\times$Temporal Attention in Appendix~\ref{Sec:architecture}. We also provide more details on loss function, training degradations and inference strategies of our model in Appendix~\ref{Sec:training}. Furthermore, we provide further ablation studies on training strategies and parameter settings in Appendix~\ref{Sec:more_ablation}. Finally, we include extensive comparisons on temporal consistency, and we provide more comparisons with state-of-the-art diffusion-based VSR models and additional qualitative results to demonstrate the effectiveness of our approach in Appendix~\ref{Sec:more_results}.

\section{Architecture}
\label{Sec:architecture}

\subsection{Architecture Details}
We follow the original ViViT~\cite{vivit} architecture and adapt it to meet the requirements of the VSR task. Tab.~\ref{tab:inputembedding_hyperparameters},\ref{tab:dualxtransformer_hyperparameters} and \ref{tab:reconstruction_hyperparameters} provide an overview of the hyper-parameters for the Input Video Embedding, Dual Axial Spatial$\times$Temporal Transformer, and Spatiotemporal Reconstruction Module, respectively. Our code and pre-trained models (both MSE and GAN models) will be made publicly available.

\begin{table}[htbp]
\centering
\resizebox{0.35\textwidth}{!}{
\begin{tabular}{l c}
\toprule[1.5pt] 
\textbf{Hyper-parameter}     & \textit{Input Video Embedding} \\ \hline
Image patch size                         & $2\times2$           \\ 
Frame patch size                         & $1$           \\  \hline
CNN kernel size             & $3,3$ \\
CNN channels                    & 64          \\ \hline
Temporal Attention depth  & 2 \\
Embedding dimension             & 64 \\
MLP dimension           &  128 \\ 
Head number           &  12 \\ 
Positional encoding   & RoPE~\cite{rope} \\ \toprule[1.5pt]
\end{tabular}
}
\caption{Hyper-parameter configuration for Input Video Embedding, including patchification settings and architecture details, similarly for the following tables.}
\label{tab:inputembedding_hyperparameters}
\end{table}

\begin{table}[htbp]
\centering
\resizebox{0.4\textwidth}{!}{
\begin{tabular}{l c}
\toprule[1.5pt] 
\textbf{Hyper-parameter}     & \textit{Dual Axial Spatial$\times$Temporal Transformer} \\ \hline
Vertical-Temporal Attention depth  & 6 \\
Horizontal-Temporal Attention depth  & 6 \\
Embedding dimension             & 1280 \\
MLP dimension           &  2560 \\ 
Head number           &  12 \\ 
Positional encoding   & RoPE \\
\toprule[1.5pt]
\end{tabular}
}
\caption{Hyper-parameter configuration for Dual Axial Spatial$\times$Temporal Transformer.}
\label{tab:dualxtransformer_hyperparameters}
\end{table}

\begin{table}[t]
\centering
\resizebox{0.4\textwidth}{!}{
\begin{tabular}{l c}
\toprule[1.5pt] 
\textbf{Hyper-parameter}     & \textit{Spatiotemporal Reconstruction Module} \\ \hline
Temporal Attention depth  & 2 \\
Embedding dimension             & 1280 \\
MLP dimension           &  2560 \\ 
Head number           &  12  \\ 
Positional encoding   & RoPE \\ \hline
CNN kernel size             & $3,3$ \\
CNN channels                    & 64 \\ \hline
Pixel shuffle scale factor     & 2 \\
\toprule[1.5pt]
\end{tabular}
}
\caption{Hyper-parameter configuration for Spatiotemporal Reconstruction Module.}
\label{tab:reconstruction_hyperparameters}
\end{table}


\subsection{Further Discussion of Dual Axial Spatial$\times$Temporal Attention}

\noindent\textbf{Attention Modules with Horizontal and Vertical Directions.} Previous low-level vision models also include attention modules with horizontal and vertical directions, such as Stripformer~\cite{stripformer}, which is used for image deblurring. Stripformer's attention mechanism combines intra-strip and inter-strip attention to capture blur patterns, with horizontal and vertical computations similar to DualX-VSR. However, while Stripformer targets fast image deblurring, DualX-VSR is designed for Real-World VSR. Additionally, DualX-VSR implements \textbf{\underline{3D}} Horizontal-Temporal and Vertical-Temporal attention, respectively, unlike Stripformer's \textbf{\underline{2D}} spatial attention. To the best of our knowledge, DualX-VSR is the first to introduce 3D dual axial spatial$\times$temporal attention in the real-world video super-resolution task.

\noindent\textbf{Spatial-Temporal Attention.} As discussed in Sec. 3.1 of the main paper, methods like RVRT~\cite{rvrt}, PSRT~\cite{psrt}, and IART~\cite{iart} use local window-based attention to fuse spatiotemporal information, treating spatiotemporal components \textbf{\underline{separately}}, which limits the receptive field and scalability. In contrast, DualX-VSR, as described in Sec. 3.2 of the main paper, treats spatiotemporal information \textbf{\underline{holistically}}, fusing it across patch, token, and attention levels, offering a more video-specific approach.

\section{Details on Training and Inference}
\label{Sec:training}

\subsection{Loss Function}
For the output fidelity loss $\mathcal{L}_{\text{pix}}$ we use Charbonnier loss~\cite{charbonnier}. In addition, we use perceptual loss~\cite{preceptual} $\mathcal{L}_{\text{per}}$ and adversarial loss~\cite{adversarial} $\mathcal{L}_{\text{adv}}$ to achieve better visual quality.

Our training process is divided into three stages: pretraining on bicubic downsampling, finetuning with degradations, and further finetuning with the addition of perceptual loss and adversarial loss. In the first and second stages, we train DualX-VSR with only the fidelity loss:
\begin{equation}
\mathcal{L}_{\text{1st}} = \mathcal{L}_{\text{2nd}} = \mathcal{L}_{\text{pix}}, 
\end{equation}

In the third stage, we finetune the network with additional perceptual loss and adversarial loss:
\begin{equation}
\mathcal{L}_{\text{3rd}} = \lambda_{\text{pix}} \mathcal{L}_{\text{pix}} +  \lambda_{\text{per}} \mathcal{L}_{\text{per}} + \lambda_{\text{adv}} \mathcal{L}_{\text{adv}}.
\end{equation}

In our experiments, $\lambda_{\text{pix}}=1\times 10^{-2}$, $\lambda_{\text{per}}=1$ and $\lambda_{\text{adv}}=5 \times 10^{-3}$ following BasicVSR~\cite{basicvsr}.

\subsection{Training Degradations}
\begin{figure}[t]
  \centering
   \includegraphics[width=0.8\linewidth]{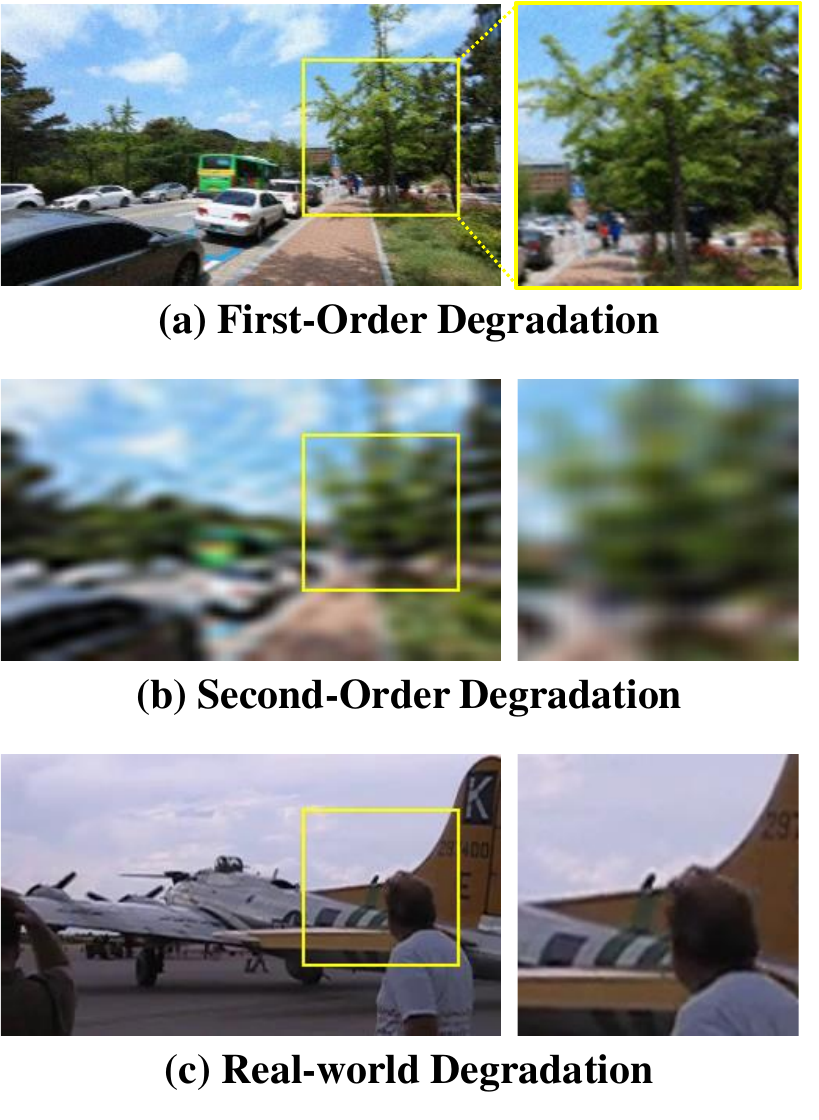}
   \caption{Comparison of different degradations. (a) and (b) are obtained by applying first-order and second-order degradations on videos from the REDS4~\cite{reds} dataset respectively, while (c) is sampled from real-world videos in the VideoLQ~\cite{realbasicvsr} dataset. As observed, the first-order degradation in (a) closely resembles the real-world degradation in (c), while the second-order degradation in (b) deviates significantly from real-world degradation.}
   \label{fig:degradation}
\end{figure}

Following RealBasicVSR~\cite{realbasicvsr}, we adopt a similar degradation synthesis pipeline. Specifically, we apply random blur, resize, noise, and JPEG compression as image-based degradations following Real-ESRGAN~\cite{realesrgan}. We also apply video compression following RealBasicVSR. Unlike RealBasicVSR, we constrain the degradation parameters to a first-order model. This is based on our experimental findings that first-order degradation is sufficient to approximate real-world low-quality videos. In contrast, second-order degradation tends to produce overly severe artifacts, making it less representative of real-world scenarios. The comparison of different degradations is shown in Fig.~\ref{fig:degradation}.

Since this pipeline involves random parameters, PSNR values may vary from those in other papers. For fairness in comparison, we ensure consistent degradation across datasets. All compared methods were trained using the same degradation, avoiding out-of-distribution issues.

\subsection{Inference Strategies}
Following previous works~\cite{iart,rvrt,psrt}, we set the input patch size to $112\times112$ for DualX-VSR (MSE) and $64\times64$ for DualX-VSR (GAN) during inference, consistent with the training setting. The number of input frames is fixed to 16, matching the training configuration. Furthermore, to ensure temporal and spatial consistency, we apply an overlapping strategy during inference in both temporal and spatial dimensions, inspired by the approaches in \cite{vrt,iart,rvrt,psrt}.

\section{More Ablation Studies}
\label{Sec:more_ablation}

We conduct more ablation studies to assess the impact of various settings, including pre-training strategy, train patch size, and train frame number. The experimental results are shown in Tab.~\ref{tab:ablation}, where (d) represents DualX-VSR (MSE). 

\begin{table}[t]
\centering
\resizebox{0.48\textwidth}{!}{%
\begin{tabular}{c | c c c | c c}
\toprule[1.5pt] 
Exp. & Pre-training & Train Patch Size & Train Frame Number & PSNR $\uparrow$ & SSIM $\uparrow$ \\ \hline
(a) & $\times$ & $112\times 112$ & 16 & 25.71 & 0.6973 \\ 
(b) & \checkmark & $64\times64$  & 16 & 25.84 & 0.7021 \\ 
(c) & \checkmark & $112\times 112$ & 8  & 25.85 & 0.7022 \\ 
\rowcolor{gray!15}(d) & \checkmark & $112\times 112$ & 16 & \textbf{26.03} & \textbf{0.7092} \\ 
\bottomrule[1.5pt] 
\end{tabular}%
}
\caption{More ablation studies of DualX-VSR. Traning setting of DualX-VSR (MSE) are marked with \colorbox{gray!15}{gray} background. The best performances are marked with \textbf{bold}.}
\label{tab:ablation}
\end{table}

\noindent\textbf{Effectiveness of Pre-training.} DualX-VSR adopts a pretrain-finetune strategy, where it is first pre-trained on a dataset with only bicubic downsampling, followed by fine-tuning on a dataset with degradations. As shown in Tab.~\ref{tab:ablation}, the performance of (a) which lacks pre-training drops significantly, indicating the effectiveness of pre-training.

\noindent\textbf{Train Patch Size \& Frame Number.} In the pre-training stage, DualX-VSR follows the same settings like other VSR methods~\cite{realbasicvsr,realviformer}, with a train frame number of 16 and a train patch size of $64\times64$. During fine-tuning on the dataset synthesized with degradation, we increased the train patch size to $112\times112$ while keeping the train frame number unchanged to provide a larger receptive field that helps the model better adapt to various degradations. The comparison of (b) and (d) in Tab.~\ref{tab:ablation} confirm that increasing the train patch size effectively enhances the model's learning capability under degradation. Comparison of (c) and (d) demonstrates that increasing the train frame number also significantly improves model performance, further illustrating that our model adopts a global perspective when modeling videos and benefits from a larger receptive field.

\begin{figure*}[t]
  \centering
   \includegraphics[width=0.95\linewidth]{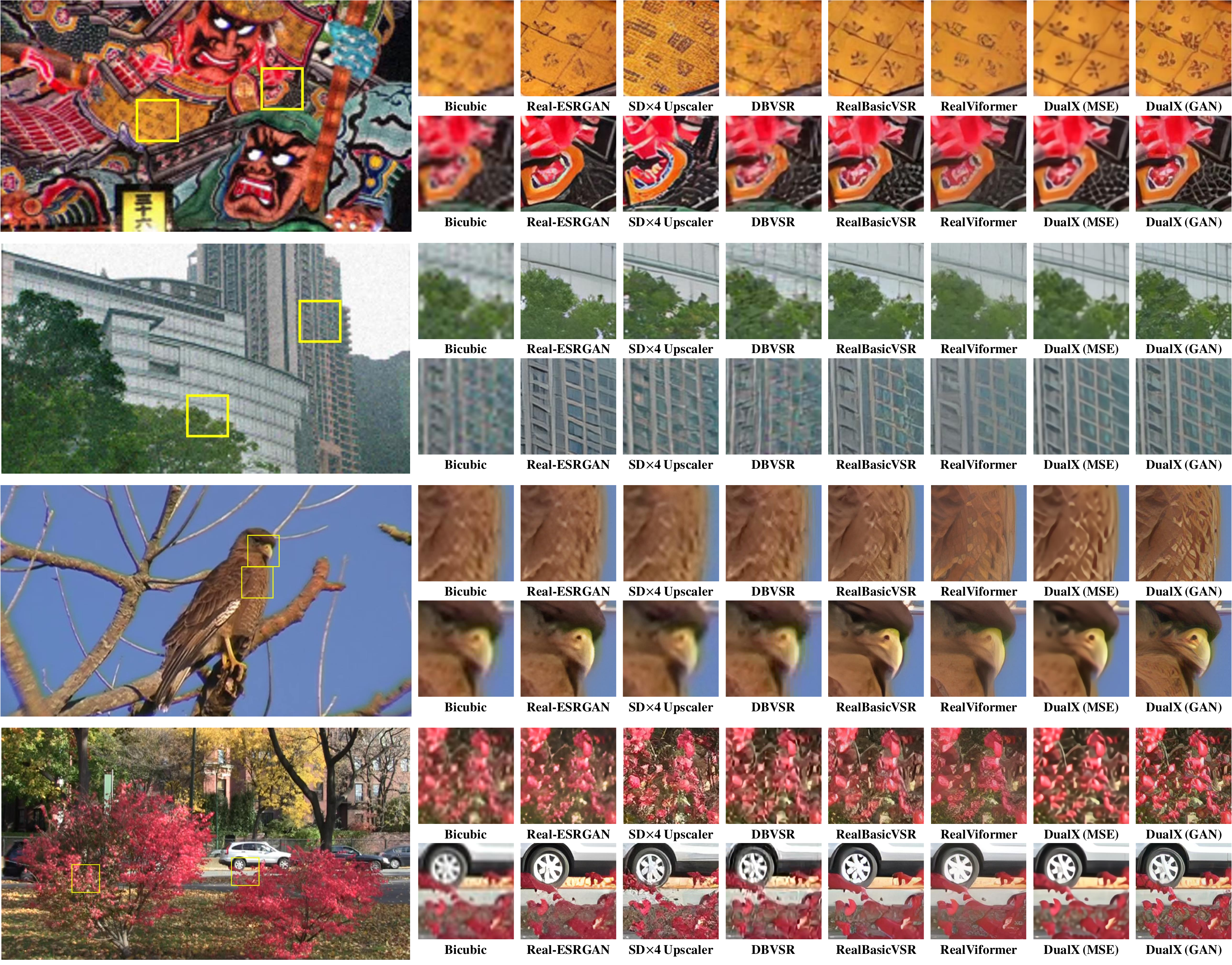}
   \caption{More qualitative comparisons on both synthetic datasets and real-world datasets. The top two comparisons are from the UDM10 and SPMC30~\cite{spmc}, while the bottom two are from VideoLQ. Our DualX-VSR (MSE) and DualX-VSR (GAN) demonstrate excellent visual performance on both synthetic and real-world datasets, effectively removing degradations and achieving high-fidelity results.}
   \label{fig:comparison_ablation}
\end{figure*}

\begin{figure}[t]
  \centering
   \includegraphics[width=0.99\linewidth]{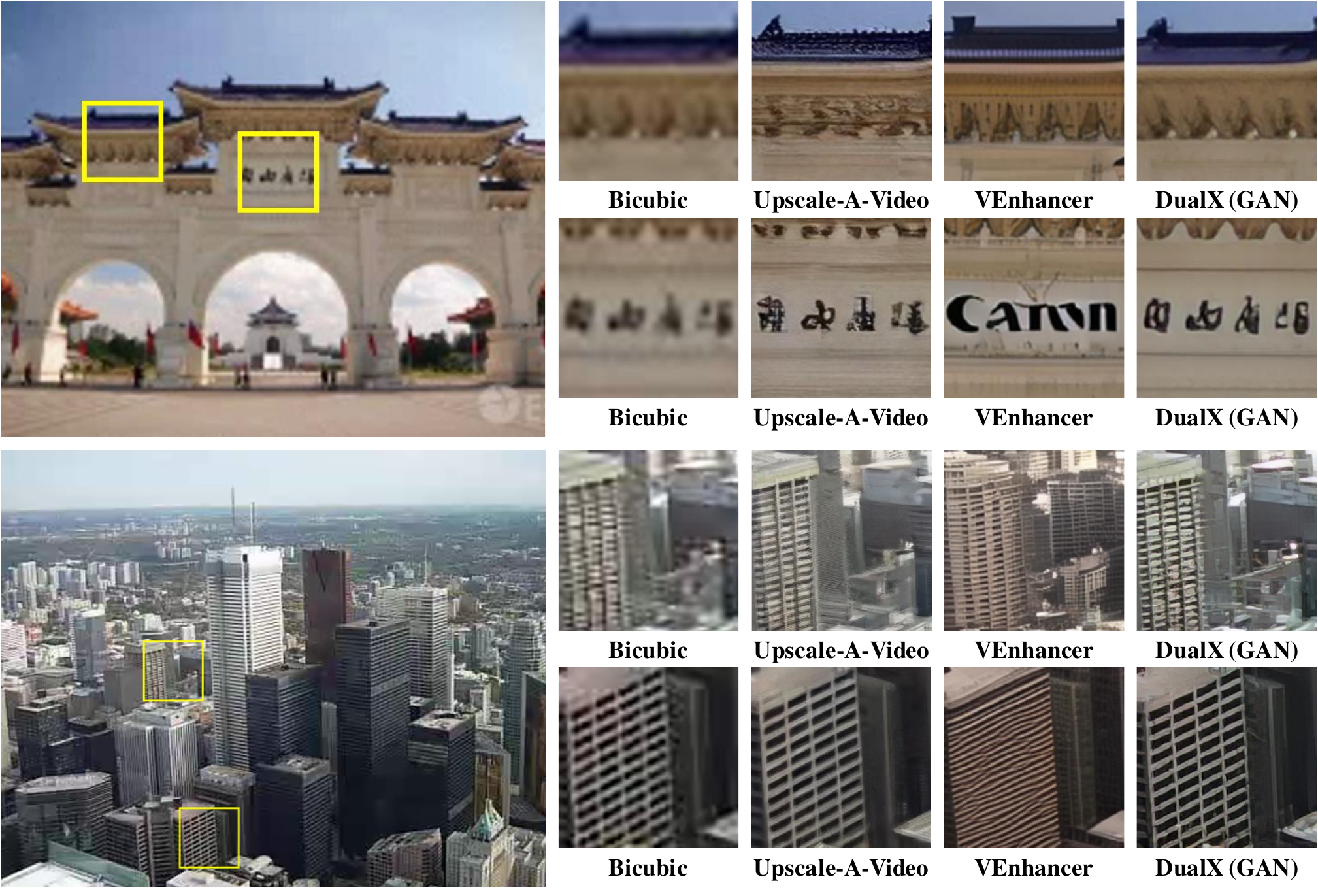}
   \caption{Qualitative comparison with diffusion-based real-world VSR models. Diffusion-based models often produce overly creative generation results, leading to unrealistic textures. In contrast, our DualX-VSR (GAN) model can restore the real details of videos more faithfully, achieving better fidelity.}
   \label{fig:compare_diffusion}
\end{figure}

\begin{table}[htbp]
    \centering
    \resizebox{0.4\textwidth}{!}{%
    \begin{tabular}{c c c}
    \toprule[1.5pt] 
         Model  &  $E_{warp}^*$\cite{ewarp}$\downarrow$/BC\cite{vbench,vbench++}$\uparrow$ \\ \hline
        
        RealVSR\cite{realvsr} & 1.570 / 0.958 \\ 
        RealBasicVSR\cite{realbasicvsr} & 0.541 / \textcolor{blue}{\textbf{0.960}} \\ 
        RealViformer\cite{realviformer} & \textcolor{red}{\textbf{0.421}} / 0.959  \\ 
        Upscale-A-Video\cite{upscaleavideo} & 0.658 / 0.958 \\ 
        \rowcolor{gray!15}\textbf{DualX-VSR(GAN)} & \textcolor{blue}{\textbf{0.532}} / \textcolor{red}{\textbf{0.961}} \\ 
        \bottomrule[1.5pt] 
    \end{tabular}
    }
    \caption{Comparison of temporal consistency on UDM10 dataset. The best and second-best performances are highlighted in \textcolor{red}{\textbf{red}} and \textcolor{blue}{\textbf{blue}}, respectively. $E_{warp}^*$ denotes $E_{warp}(\times 10^{-3})$. BC refers to BackgroundConsistency in VBench~\cite{vbench,vbench++}.}
    \label{tab:complexity_temporal_supp}
\end{table}

\section{More Results}
\label{Sec:more_results}

\subsection{More Comparisons on Temporal Consistency}
We provide additional comparisons of temporal consistency metrics, as shown in Tab.~\ref{tab:complexity_temporal_supp}. On the UDM10~\cite{udm} dataset, DualX-VSR also demonstrates competitive temporal consistency performance over other methods. It is important to note that there exists a trade-off between temporal consistency and single-frame image quality. Some models often compromise visual quality in order to achieve improved continuity. In contrast, DualX-VSR attempts to strike a balance between these two aspects, simultaneously achieving better image quality and temporal consistency.

\subsection{Qualitative Comparisons with Diffusion-based Real-world VSR Models}
We compare DualX-VSR with current state-of-the-art diffusion-based real-world VSR models, including Upscale-A-Video~\cite{upscaleavideo} and VEnhancer~\cite{venhancer}. For a fair comparison, both models were evaluated with no prompt. As shown in Tab.2 in main paper, DualX-VSR (GAN) achieves the best performance on synthetic datasets REDS4~\cite{reds} and UDM10~\cite{udm} in terms of metrics including PSNR, SSIM, LPIPS~\cite{psnr}. Additionally, it outperforms on the real-world dataset VideoLQ~\cite{realbasicvsr} in terms of NIQE~\cite{niqe} and LIQE~\cite{liqe}, highlighting the high-fidelity results of our approach. Qualitative comparisons in Fig.~\ref{fig:compare_diffusion} further show that our model reconstructs real-world details more faithfully than diffusion-based models.


\subsection{More Qualitative Comparisons}
We present additional qualitative comparisons with state-of-the-art image and video super-resolution models across both synthetic and real datasets in Fig.~\ref{fig:comparison_ablation}, where DualX-VSR also demonstrates great performance.